\documentclass[5p,times]{elsarticle}
\usepackage[colorlinks=false,hidelinks]{hyperref}
\usepackage{lineno}
\usepackage[usenames,dvipsnames]{xcolor}
\usepackage{mathtools}
\usepackage{bbold}
\usepackage{siunitx}
\usepackage{amssymb}
\usepackage{cases}
\usepackage{algorithm}
\usepackage{algpseudocode}
\usepackage{color,soul}
\newcommand{\argmin}[1]{\underset{#1}{\operatorname{arg}\,\operatorname{min}}\;}
\newcommand{\norm}[1]{\ensuremath{\lVert #1 \rVert}}
\newcommand{\minimize}[1]{\underset{#1}{\operatorname{minimize}}\;}
\DeclarePairedDelimiter{\normal}{\lVert}{\rVert}
\makeatletter
\def\ps@pprintTitle{%
	\let\@oddhead\@empty
	\let\@evenhead\@empty
	\let\@oddfoot\@empty
	\let\@evenfoot\@oddfoot
}
\makeatother
\journal{Arxiv}

\usepackage{framed} 
\usepackage{nomencl} 
\makenomenclature

\begin{document}

\begin{frontmatter}
	\title{An Unsupervised Method for Estimating the Global Horizontal Irradiance from Photovoltaic Power Measurements}
	
	\author[epfl,isaac]{Lorenzo Nespoli \corref{mycorrespondingauthor}}
	\cortext[mycorrespondingauthor]{Corresponding author}
	\ead{lorenzo.nespoli@epfl.ch}
	\author[isaac]{Vasco Medici}
	
	\address[epfl]{Swiss Federal Institute of Technology in Lausanne, Switzerland}
	\address[isaac]{University of Applied Sciences and Arts of Southern Switzerland, Lugano, Switzerland}
	
	\begin{abstract}
		The precise calculation of solar irradiance is pivotal for forecasting the electric power generated by PV plants. However, on-ground measurements are expensive and are generally not performed for small and medium-sized PV plants. Satellite-based services represent a valid alternative to on-site measurements, but their space-time resolution is limited. In this paper we present a method for estimating the global horizontal irradiance (GHI) from the power measurements of one or more photovoltaic (PV) systems located in the same neighborhood. The method is completely unsupervised and is based on a physical model of a PV plant. It can estimate the nominal power and orientation of multiple PV fields, using only the aggregated power signal from their PV power plant. Moreover, if more than one PV power plant is available, the different signals are reconciled using outliers detection and assessing shading patterns for each PV plant. Results from two case studies located in Switzerland are presented here. The performance of the proposed method at estimating GHI is compared with that of free and commercial satellite services. Our results show that the method presented here is generally better than satellite-based services, especially at high temporal resolutions.
	\end{abstract}
	
	
\end{frontmatter}

\nomenclature{$\widehat{P}$}{estimated power (W)} 
\nomenclature{$T$}{number of time steps (-)}
\nomenclature{GHI}{global horizontal irradiance [$W/m^2$]} 
\nomenclature{$T_a$}{ambient temperature [$^\circ C$]}
\nomenclature{$\beta$}{pv field azimuth [rad]}
\nomenclature{$\alpha$}{pv field tilt [rad]}
\nomenclature{$g$}{vector of geographical coordinates}
\nomenclature{$P_{nom}$}{PV field nominal power [$W$]}
\nomenclature{GHI$^\star$}{optimized value of GHI [$W/m^2$]}
\nomenclature{$n_{pv}$}{number of PV power signals [$-$]}
\nomenclature{$P^n$}{normalized power signal [$-$]}
\nomenclature{$\mathbf{e}$}{PV estimation error [$W$]}
\nomenclature{$\mathcal{F}$}{trust function}
\nomenclature{$h$}{objective function}
\nomenclature{$\gamma_s$}{sun azimuth [rad]}
\nomenclature{$\theta_z$}{sun zenith [rad]}
\nomenclature{$\mathcal{N}$}{normal distribution}
\nomenclature{$\mu$}{mean of distribution } 
\nomenclature{$\sigma$}{standard deviation of the distribution}
\nomenclature{$\mathbf{Pr}$}{proxy matrix [$W/m^2$]}
\nomenclature{$\mathbf{\omega}$}{coefficients related to $\mathbf{Pr}$ [$m^2$]}
\nomenclature{$\alpha_{pr}$}{proxy tilt [rad]}
\nomenclature{$\beta_{pr}$}{proxy azimuth [rad]}
\nomenclature{DHI}{diffuse horizontal irradiance [$W/m^2$]}
\nomenclature{DNI}{direct normal irradiance [$W/m^2$]}
\nomenclature{$\mathrm{AOI}$}{angle of incidence [rad]}
\nomenclature{$\mathrm{I}$}{irradiance on an oriented surface [$W/m2$]}
\nomenclature{$\mathrm{I}_b$}{beam component of the irradiance on an oriented surface [$W/m2$]}
\nomenclature{$\mathrm{I}_d$}{diffuse irradiance on an oriented surface [$W/m2$]}
\nomenclature{$\mathrm{I}_g$}{ground reflected irradiance on an oriented surface [$W/m2$]}
\nomenclature{$\rho$}{albedo [$-$]}
\nomenclature{$\mathrm{IAM}$}{incident angle modifier [$-$]}
\nomenclature{$T_{cell}$}{PV cell temperature [$^\circ C$]}
\nomenclature{$T_{ref}$}{reference temperature [$^\circ C$]}
\nomenclature{$\phi$}{PV cell temperature correction coefficient [$Km^2/W$]}
\nomenclature{$\gamma$}{PV power temperature coefficient [1/K]}
\nomenclature{$\mathrm{I}^\mathrm{AOI}$}{irradiance corrected with the angle of incidence [$W/m^2$]}
\nomenclature{$\mathrm{I}^\mathrm{AOIT}$}{irradiance corrected with the angle of incidence and temperature [$W/m^2$]}
\nomenclature{$\eta$}{combined module and inverter efficiency [$-$]}
\nomenclature{$\mathrm{I}_{\mathrm{STC}}$}{reference irradiance [$W/m^2$]}
\nomenclature{$\delta$}{fixed GHI increment for the computation of derivatives [$W/m^2$]}
\nomenclature{$\nu$}{current iteration [$-$]}
\nomenclature{$\Vert \cdot \Vert_{fr}$}{Frobenius norm }
\nomenclature{$\lambda$}{stepsize for the optimization algorithm [$-$]}
\nomenclature{$\mathrm{re}_{clear}$}{relative PV estimation error with respect to clear sky condition[$-$]}
\nomenclature{$\mathcal{I}$}{outlier detection function}
\nomenclature{$\mathbf{E}$}{estimation error matrix [$W$]}
\nomenclature{$\mathcal{L}$}{robust loss function} 
\nomenclature{$\mathrm{Pr}$}{proxy [$W/m^2$]}
\nomenclature{$\omega$}{coefficient associated to Pr [$m^2$]}

\section{Introduction}
\subsection{Motivation}
Solar power generation, both at utility and residential level, will play a central role in the future of the electric power industry, with a predicted installed power ranging from 4.3 to 14.8TW by 2050 \cite{mayer2015current}. Although this trend is certainly to be welcomed, unless countermeasures are taken the intermittent nature of solar generation could lead to stability issues in the electrical grid \cite{ eftekharnejad2013impact}. In the distribution grid, these problems will be further emphasized by the increase of electricity consumption driven by the electrification of heat generation and mobility \cite{IEA2016}, which will further increase the amplitude of the power and voltage fluctuations \cite{richardson2010impact}.\\
Fortunately, in the meanwhile smart grid solutions that help to overcome the above-mentioned issues by modulating generation and demand are becoming available and affordable. For example, distributed energy storage systems \cite{bahramipanah2016decentralized} and demand side management \cite{gelazanskas2014demand,Kim2013} can be exploited for the active control of distribution networks.\\
\begin{framed}
	\printnomenclature				
\end{framed}
To optimize the control and guarantee the quality-of-service in the electricity grid, it is important to predict the power flows accurately, as this favors a sensible management of the available flexibility. It has been shown that forecasting accuracy can be improved when the production and consumption in the grid are disaggregated and predicted separately \cite{Monforte,Shaker2014}. The disaggregation of solar generation from the total grid load can be achieved by using on ground irradiance measurements \cite{Sossanb}. In general, global horizontal irradiance (GHI) measurements are often used as exogenous input when performing both long term and short term PV production forecasts \cite{lorenz2009irradiance,Inman2013}. Unfortunately, accurate on ground irradiance measurements are often not available. Although irradiance measurements can also be used for the online estimation of PV power production and for fault detection, sensors are usually not installed for small and medium-sized plants, due to their high cost. If the irradiance is not measured directly by means of on-ground measurements, satellite estimations can be exploited. Satellite-based radiation assessment services provide an estimate of the time course of GHI for a given location, but their spatio-temporal resolution is constrained by technical limitations. Most of these services are based on the images acquired by the Meteosat 2nd generation satellites, which have a spatial resolution of 3 km at the nadir and a temporal resolution of 15 minutes \cite{Mecikalski2007}. These coarse resolutions limit the performance of satellite-based nowcasting methods. Moreover, the limited spatial resolution has a smoothing effect that can result in reduced accuracy levels for GHI estimation at a specific location, especially in the presence of local clouds. The active control of distribution networks, of which some of the critical sections can take up a small area, requires a more accurate and fast estimate of GHI. Satellite-based methods could also profit from an increased availability of on-ground GHI measurements, as they could be used for calibration \cite{Vernay2013,Mieslinger2014,Lorenzo2017}, a technique also known as site adaptation.\\
In this study, we investigate the possibility of using local PV power measurements to estimate GHI with a high temporal and spatial accuracy. 
Being able to estimate GHI directly from PV power measurements will allow to better estimate and forecast the PV production of an entire neighborhood by monitoring the power output of only a small fraction of the PV plants, without the need to install additional irradiance sensors.\\
For the development of the proposed method, we focused on accessibility and simplicity. Indeed, the method is fully unsupervised and the only inputs it requires are the measurements of the AC power output of the monitored PV plants, the ambient temperature and the geographical coordinates of the neighborhood.

  
\subsection{Previous work}\label{previous_work}
The idea of using PV plants as surrogated irradiation sensors has already been researched in the past. In \cite{Laudani2016}, voltage and current measurements from a PV module were used to calculate  the incoming radiation on the plane of array. In \cite{Kara2016}, it is suggested that a nearby PV plant can be used as a proxy to estimate the power generated by another PV installation, even if only a linear relationship is considered between the proxy and the estimated output.
In \cite{Traxler}, proximate PV plants are used to predict the PV outputs of other PV installations, for the purposes of automatic  fault detection. 
In \cite{Engerer2014}, a clear sky index, $K_{pv}$, is introduced: this is the ratio between the AC power of a simulated PV plant under clear sky conditions and the actual power measurements. The authors use a clear-sky radiation model, a transposition model and an inverter and PV module model. This method relies on an accurate, technical description of the PV system, which includes the PV module orientation. In \cite{Killinger2016}, a methodology is proposed to project power generation between different PV systems. The PV system power output is modeled as a quadratic function of the solar irradiance on the plane of array (POA) and the ambient temperature. The five coefficients of the curve must be  fitted  for each type of PV module technology. The POA irradiance is obtained by inverting the quadratic expression. The GHI is then estimated from the POA irradiance, using an iterative procedure. In their discussion, the authors suggest that simultaneously considering PV systems with multiple orientations could increase the accuracy of the GHI estimation. In \cite{Marion2017}, a similar methodology is presented for obtaining GHI from PV power measurements. Building on the work of \cite{Killinger2016}, a correction for low angle of incidence and wind speed is considered. The AOI correction is based on 18 coefficients, specific to the type of PV module coating considered. In all the aforementioned studies, namely \cite{Engerer2014,Killinger2016,Marion2017}, it is assumed that the inverter type, PV module type and PV module orientation are known. Furthermore, it is assumed that all the modules of a given PV plant have the same orientation.

\subsection{Outline and objective}
Despite the increasing number of PV installations and the abundance of available monitoring data, it is difficult to use them to estimate the GHI signal. Considering the works previously cited in subsection \ref{previous_work}, we can identify three main causes that make this task particularly challenging:
\begin{enumerate}
	\item Most of the estimation methods in the literature require a detailed description of the PV systems, including PV power plant nominal power, fields orientations, module and inverter types.  Gathering all these metadata is difficult and time consuming. Moreover, when available, the data contained in databases could be imprecise or flawed \cite{Killinger2017}. This could lead to erroneous estimations.
	\item  Occasionally, PV power plants can be composed of different PV fields, each field having different nominal power and orientations. A typical case is a PV power plant with an east-west configuration, but more complex configurations are possible, as presented later in the paper. In this cases, if we possess only a single power signal, we should be able to retrieve nominal powers and orientations of an arbitrary number of PV fields to correctly estimate GHI.
	\item If more than a power signal is available for a given geographical location, an automated procedure is needed to reconcile all the measurements and efficiently make use of all the signals.
\end{enumerate}
We present here a fully unsupervised method for estimating the local GHI using only the power measurements from one or more PV plants, without the need to know their nominal power and module orientations. The problem of identifying PV plants with differently oriented modules from a single power signal is addressed by means of a robust regression. In order to increase the accuracy of the GHI estimation, the method can exploit multiple power signals from different PV plants. 
The different signals are reconciled  by means of outlier detection and by determining  the shading patterns of each PV plant. The code related to the GHI estimation method, including the PV system identification methodology, is freely available online (see Section 7).

The paper is structured as follows: Section \ref{sec:methodology} describes the methodology used for estimating the GHI. Section \ref{sec:orientation} discusses the issue of how to identify the PV plant orientations without knowing the actual GHI. Section \ref{sec:proxy} discusses the models used to obtain the PV power production proxies. Section \ref{sec:numerical} briefly discusses the numerical methods for solving the GHI estimation problem. In Section \ref{sec:case}, the accuracy of the method is assessed for two case studies, and compared to satellite-based GHI estimations and secondary standard pyranometer measurements. Finally, conclusions are presented in Section \ref{sec:conclusions}.

\section{Methodology}\label{sec:methodology}
The combined effect of irradiance and ambient temperature on the PV power production is well-known. Accurate empirical models that assess the total incoming irradiation on an oriented surface, given the GHI, are also available \cite{yang2016,Loutzenhiser2007}.
We can therefore build a function that links the GHI to a given PV plant power output:

\begin{equation}\label{eq:p_of_ghi}
\mathbf{\widehat{P}} = f(\mathbf{GHI},\mathbf{t},\boldsymbol{\alpha},\boldsymbol{\beta},\mathbf{g},\mathbf{T}_a,\mathbf{P_{nom}})
\end{equation}
where $\mathbf{\widehat{P}} \in {\rm I\!R}^{T \times 1}$ is the estimate of the power generated by a given PV plant, where $T$ is the number of time steps in the data, $\mathbf{GHI} \in {\rm I\!R}^{T \times 1}$ and $\mathbf{T}_a \in {\rm I\!R}^{T \times 1}$ are the vectors of the observed GHI and temperatures at times $\mathbf{t}$ $\in {\rm I\!R}^{T \times 1}$, $\boldsymbol{\alpha}$ and $\boldsymbol{\beta}$ are the vectors containing the tilts and azimuths of the modules, $\mathbf{g}$ is a vector containing the geographical coordinates of the plant, namely latitude, longitude and elevation, and $\mathbf{P_{nom}}$ is the vector of the nominal powers of the modules.
Function $f$ is described later by equations \ref{eq:DHI}-\ref{eq:t} and by the empirical disc model, as stated in Section \ref{sec:proxy}.
If the module orientations and nominal powers were known, $f$ could be inverted in order to estimate GHI. Unfortunately, $f$ is not always invertible, especially when $\mathbf{\widehat{P}}$ comes from a PV plant with differently oriented PV modules.
So, for different values of GHI, the function $f$ could return the same output $P$. In this case, a method is required in order to choose the correct GHI value from a range of possible choices. This problem is solved by following two steps. First, we estimate the panel orientation from the measured power of the given PV plant. We then use the calculated orientations to build function $f$ and solve 

\begin{equation}\label{eq:master}
\mathbf{GHI^\star}  = \argmin{\mathbf{GHI}} \norm{\mathbf{P}-\widehat{\mathbf{P}}}_2
\end{equation}
without inverting $f$. Here $\mathbf{GHI^\star}$ refers to the optimized values of $\mathrm{GHI}$. \\
Equation \ref{eq:master} can be solved by using one or more PV power signals and can therefore be easily reformulated as:
\begin{equation}\label{eq:master2}
\mathbf{GHI^\star}  = \argmin{\mathbf{GHI}}  \normal*{\frac{1}{n_{pv}} \sum_{i=1}^{n_{pv}} \mathbf{P}^n_i-\widehat{\mathbf{P}}_i^n}_2
\end{equation}
where $n_{pv}$ is the total number of PV power signals and $\mathbf{P}^n_i$ and $\widehat{\mathbf{P}}_i^n$ are the observed and estimated power signals normalized with the estimated nominal power (see Section \ref{sec:orientation} for how this is estimated). 
The main drawback to this formula is that, when estimating $\mathrm{GHI}$, all the PV signals are equally weighted. This solution is not robust in the event of shadows or faulty signals. 
Equation \ref{eq:master2} can be improved in two ways:

\begin{enumerate}
	\item Faulty signals can be statistically detected, to avoid using them for estimating $\mathrm{GHI}$
	\item When the modules are partially or completely shaded, the GHI estimation is not accurate. This effect can be mitigated by building a map of the error $\mathbf{e}_i = \mathbf{P}^n_i-\widehat{\mathbf{P}}^n_i$, as a function of the sun position (azimuth and elevation). This map can then be used to evaluate how much a certain PV measurement can be trusted as a function of sun position. For the rest of the paper, we will refer to this map as the trust function.
\end{enumerate}
The above considerations lead to the more general formula:
\begin{equation}\label{eq:master3}
\mathbf{GHI^\star}  = \argmin{\mathbf{GHI}}  \normal*{\frac{1}{n_{pv}}  \sum_{i=1}^{n_{pv}} \mathcal{I}(\mathbf{E}) \mathcal{F}_i(\boldsymbol{\gamma}_s,\boldsymbol{\theta}_{z}) \mathbf{e}_i}_2
\end{equation}
where $\mathcal{I}:\ {\rm I\!R} \rightarrow \left\lbrace0,1\right\rbrace $ is an indicator function detecting the presence of outliers, $\mathbf{E} = [\mathbf{e}_1,\mathbf{e}_2,...\mathbf{e}_N]$ is the estimation error matrix, $\mathcal{F}:\ {\rm I\!R} \rightarrow [0,1] $ is the trust function and $\boldsymbol{\gamma}_s$ and $\mathbf{\theta}_{z}$ are, respectively, the azimuth and zenith of the sun.
Here $\mathcal{F}$ can be interpreted as a dynamic weight function, since $\boldsymbol{\gamma}_s$ and $\mathbf{\theta}_{z}$ are function of time. The role of the trust function is to attach less importance to the calculation of the ith signal if this is believed to be affected by shadows with a particular position of the sun.
The construction of the $\mathcal{I}$ and $\mathcal{F}$ functions is described in Section \ref{sec:numerical}.

\section{Orientation assessment}\label{sec:orientation}
PV plant orientation could theoretically be estimated from the PV plant power measurements, and from the GHI measurements, by means of the equations \ref{eq:DHI}, \ref{eq:I}, \ref{eq:I_g} and \ref{eq:I_b}. The orientation estimation can be formulated as the following optimization problem:
\begin{equation}\label{eq:estimation}
\minimize{\alpha_{i},\beta_{i}} \norm{\mathbf{P}_i-\widehat{\mathbf{P}}_i}_2
\end{equation}
where $\widehat{\mathbf{P}}_i$ is the PV production estimated from the GHI signal.
The following aspects must be taken into account:

\begin{enumerate}
	\item We want to estimate PV plant orientation without knowing the actual GHI seen by the modules 
	\item PV plants can consist of groups of modules with different orientations, e.g. plants with an east-west configuration
	\item The presence of shadows affects the relationships between the GHI projection on an oriented surface and the PV power output
	\item Problem \ref{eq:estimation} is non-linear and non-convex
\end{enumerate}
If the GHI seen by the modules is unknown, estimating their orientation would result in a blind identification problem \cite{Wills2011,Ohlsson2014}. We exploit the fact that we can obtain a good approximation of GHI for clear-sky periods, using a model for the extra -terrestrial irradiation and for the air mass index. In this paper, we used time series obtained from the Soda-pro CAMS McClear service\footnote{http://www.soda-pro.com/web-services/radiation/cams-mcclear}, which uses the McClear clear sky model \cite{Lef2013}.\\
We can thus identify the PV plant orientations if we can select clear-sky periods, using only the PV plant power output. Different methods can be used to exploit PV power signals in order to identify clear-sky radiation periods. In \cite{Lonij2012}, a period is considered to be clear if the measured PV power is higher than the 80$\%$ percentile of the set of measurements taken at the same time of day, during the previous 15 days. In \cite{Killinger2017} this method is combined with the clear-sky detection routine described in \cite{Reno2016}, which uses GHI observation as input and a set of 5 extraction parameters. In this paper, we first developed a selection based on the smoothed power signals: the power output of each PV plant is filtered using a second order low-pass Butterworth filter \cite{Butterworth1930}. We then considered a period to be clear if the root mean squared relative error between the original and filtered signal was lower than a threshold value.

The main drawback of this method is that its performance is influenced by three parameters, namely the threshold value, the low-cut frequency period and the length of the period, which have to be tuned. Moreover, in the event of PV curtailment, these curtailment periods can be identified as clear periods. \\
In order to overcome these issues, we developed a different method. PV power signal distribution  as a function of the sun position is typically bimodal, due to the presence of clouds during data acquisition. On the other hand, a unimodal distribution could indicate a systematic shadow for the corresponding sun position. 
In order to select clear data periods, we fit  a gaussian mixture probability density function with two components $\mathrm{X_i} \sim \mathcal{N}(\mu_i,\sigma_i)$ for each sun position, with a discretization of 5$^\circ$. Then, for each sun position we identify the observations lying in the one sigma interval of the gaussian distribution with the largest $\mu$ as clear observations. We chose to discard other values since they could have been potentially caused by cloud enhancement events (higher power) or by the presence of haze or high clouds (lower power). An example of a PV power distribution for a particular sun position is depicted in figure \ref{fig:bimodal}. \\
Despite the second model being more robust in terms of selection of clear sky periods, the task of predicting the GHI seen by PV panels through a clear sky model presents some intrinsic errors. Clear sky models are not perfect and it is not possible to guarantee that the GHI seen by the PV panels is exactly the one predicted by the clear sky models.
In order to overcome this problem and the others referred to above, instead of directly solving the optimization problem \ref{eq:estimation}, we reformulated it as a robust linear regression:
\begin{figure}[h]
	\centering
	\includegraphics[width=1\linewidth]{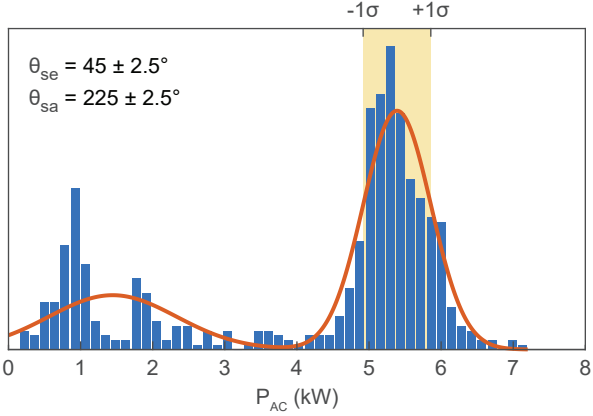}
	\caption{Distribution of the power signal of one particular PV power plant of the Biel-Benken case study, for $\theta_z = 45 \pm 2.5 ^\circ$ and $\gamma_s = 225 \pm 2.5 ^\circ$. The distribution presents a bimodal shape. We fit two gaussian distributions and considered the leftmost peak to be caused by clouds, while we select the data between one standard deviation from the mean of the rightmost gaussian distribution as belonging to clear sky periods. Values outside the one standard deviation interval were discarded, since they could have been potentially generated during cloud enhancement events (higher power) or in the presence of haze or high clouds (lower power).}
	\label{fig:bimodal}
\end{figure} 

\begin{equation}\label{eq:robust_reg}
\minimize{\boldsymbol{\omega}_i \in {\rm I\!R}^{n_p}_+ } \mathcal{L} \left(  \mathbf{P} _i -\mathbf{Pr} \boldsymbol{\omega}	_i \right)
\end{equation}
where $\mathcal{L}$ is a robust loss function \cite{Huber2009}, $\mathbf{Pr} \in {\rm I\!R}^{T \times n_p }$ is the proxy matrix, each proxy being the estimated power produced by a panel with a given orientation, $\boldsymbol{\omega}_i \in {\rm I\!R}_+^{n_p}$ is the coefficient vector for the ith PV installation, $n_p$ and $T$ being the number of proxies and total number of temporal observations. 
The additional requirement $\boldsymbol{\omega}_i \in {\rm I\!R}_+^{n_p} $ forces vector $\boldsymbol{\omega}_i$ to have all positive values.  
Thus, we can interpret the components of vector $\boldsymbol{\omega}_i$ as coefficients  describing the significance of each proxy in explaining the power output of the ith PV plant, rescaled for its nominal power. Another interpretation is that non-zero entries of $\boldsymbol{\omega}_i$ are the estimates of the nominal power of the ith PV plant oriented as its corresponding proxies.
Furthermore, problem  \ref{eq:robust_reg} forces $\boldsymbol{\omega}_i$ to be sparse  and it is robust with respect to the presence of partial shadows. \\
If we use a loss function of an M-estimator for $\mathcal{L}$, we can solve problem $\ref{eq:robust_reg}$ using an efficient iterative reweighted least square algorithm \cite{Holland1977}.
As previously anticipated, we could use more than one PV signal to estimate $\mathrm{GHI}$. In this case, we need to identify a set of coefficients for each of the $n$ signals, and some methods to assign different weights to the estimation errors arising from the different PV signals, in order to calculate $\mathrm{GHI}$ more accurately.

\section{Proxy Model for PV Performance}\label{sec:proxy}
The proxies are an estimate of the electrical power generated by a panel with a given orientation and $\mathrm{GHI}$. In order to effectively solve problem \ref{eq:estimation}, we need to select the most representative proxy orientations.  
The tilts and azimuths of the proxies, respectively $\alpha_{pr,i}$ and $\beta_{pr,i}$, are obtained by generating a triangular mesh of an icosahedron on a unit sphere. This is later refined through subdivisions, in order to increase the number of points. In this paper, the most north-facing orientations are discarded, as shown in figure \ref{fig:identifiedProxies}.\\
The sun azimuth and elevation are calculated based on the current time and the altitude, longitude and latitude of the given location.
For this task we have used the $\texttt{pvl\_ephemeris.m}$ matlab function from the freely available Sandia National Laboratories PV Collaborative Toolbox \cite{Stein2012a}, which is based on the 1985 Grover Hughes' Engineering Astronomy course at Sandia National Laboratories.
The direct normal irradiance (DNI) is then calculated by means of the empirical disc model \cite{Maxwell1987a} . The diffuse horizontal irradiance at time t $\mathrm{DHI}_t$ is then calculated as:

\begin{equation}\label{eq:DHI}
\mathrm{DHI}_t = \mathrm{GHI}_t - \cos\left(\mathrm{\theta}_{z,t}\right)\mathrm{DNI}_t
\end{equation}
where $\mathrm{\theta}_{z,t}$ is the zenith angle of the sun at time t.
DHI is then used to estimate the projection of the diffuse radiation on the given surface $\mathrm{I}_d$, using the Hay and Davies' model \cite{Loutzenhiser2007}. The overall radiation on the given surface is then given by the sum of the diffuse, direct and ground-reflected radiation.

\begin{equation}\label{eq:I}
\mathrm{I}_{i,t} = \mathrm{I}_{b,i,t} + \mathrm{I}_{d,i,t} + \mathrm{I}_{g,i,t}
\end{equation}
where $\mathrm{I}_g$ is the ground reflected component, calculated as:
\begin{equation}\label{eq:I_g}
\mathrm{I}_{g,i,t} = \rho \mathrm{GHI}_t\frac{(1-\cos(\alpha_{i}))}{2}
\end{equation}
where $\rho$ is the albedo, which was fixed to a typical value of 0.2. The direct irradiation on the oriented surface $\mathrm{I}_{b}$ is obtained from the DNI:
\begin{equation}\label{eq:I_b}
\mathrm{I}_{b,i,t} = \mathrm{DNI}_t \cos(\mathrm{AOI}_{i,t})
\end{equation} 
where $\mathrm{AOI}_i$ is the angle of incidence of the oriented surface $i$. To calculate $\mathrm{DNI}$ and $\mathrm{I}_{d,i,t}$ we used the PV Performance Modeling Toolbox by Sandia National Laboratories \cite{Stein2012a}.
Since reflection losses can significantly increase at large $\mathrm{AOI}$ \cite{Marion2017}, we applied an $\mathrm{AOI}$ correction, independent from the module technology \cite{ValinetinSoftware2012}:

\begin{equation}\label{eq;IIAM}
\mathrm{I}^{AOI}_{i,t} = \mathrm{IAM}_{i,t}  \mathrm{I}_{b,i,t} +0.95(\mathrm{I}_{d,i,t}+\mathrm{I}_{g,i,t})
\end{equation} 
	
where $\mathrm{IAM}$ is the angle of incidence modifier. We use the following ASHRAE approximation \cite{ASHRAE}: 
\begin{equation}\label{eq:IAM}
\mathrm{IAM}_{i,t} = \max \left(1-k_1\left(\cot(\min(\mathrm{AOI}_{i,t},\pi /2))-1\right),0\right)
\end{equation} 
and $k_1$ is 0.05.  

Finally, in order to obtain a proxy for the electrical power produced by a field with the ith orientation, we apply a correction taking into account the ambient temperature and the inverter and module efficiencies. The cell temperature is first estimated from the ambient temperature, then a linear correction is applied \cite{Skoplaki2009}:
\begin{equation}\label{eq:tcorrection}
\mathrm{T}_{cell,i,t} = \mathrm{T}_{a,t} +\phi \mathrm{I}^{AOI}_{i,t}
\end{equation}	

\begin{equation}\label{eq:Icorrection}
\mathrm{I}^{\mathrm{AOIT}}_{i,t} = \mathrm{I}^{\mathrm{AOI}}_{i,t} \left[1 + \gamma(\mathrm{T}_{cell,i,t}-T_{ref}) \right]
\end{equation}	
$T_{ref}=25^\circ C$ a reference temperature, $\phi$ and  $\gamma$ two coefficients. In this study, $\phi$ and $\gamma$ are not estimated and are set respectively to the values of 3.14e-2 $[Km^2 / W]$ and -4.3e-3 $[1/K]$, which represent crystalline silicon framed PV modules. Finally, the proxies are corrected for the module and inverter efficiencies, using the following equation:
\begin{equation}\label{eq:tbom}
\mathrm{Pr}_{i,t} = \eta_t\mathrm{I}^{\mathrm{AOIT}}_{i,t} 
\end{equation}
where $\eta_t$ is the combined module and inverter efficiency. In order to reduce the number of parameters, we modeled it as a function of the irradiance $\mathrm{I}^{AOIT}_{i,t}$ using the following equation:
\begin{equation}\label{eq:t}
\eta_t = k_2+k_3\ln(\mathrm{I}^{\mathrm{AOIT}}_{i,t}/I_{STC})+k_4 (\ln(\mathrm{I}^{\mathrm{AOIT}}_{i,t}/I_{STC}))^2
\end{equation}
where $I_{STC}=1000$W/m$^2$ is the reference irradiance and $k_2,\ k_3,\ k_4$ are free parameters.
By fitting equation \ref{eq:t} to typical inverter and polycrystalline module data, we obtained the following values: $k_2=0.942$, $k_3=-5.02$e-2, $k_4=-3.77$e-2.

\section{Numerical optimization}\label{sec:numerical}
\subsection{Algorithm description}
As stated in Section \ref{sec:methodology}, function $f$ is not always invertible. For this reason, we solve \ref{eq:master}, or \ref{eq:master3} if more than one PV signal is available. These minimizations can be naturally decomposed in time: that is, the norm operator can be written as the sum of the objective functions related to a single observation in time. To keep the discussion general, we can restate the left-hand side of equations \ref{eq:master}, \ref{eq:master2} and \ref{eq:master3} as:   
\begin{equation}\label{eq:decomp}
\argmin{\mathbf{GHI}} h(\mathbf{GHI})
\end{equation}
where $h$ is a placeholder for one of the objective functions defined in equations \ref{eq:master}, \ref{eq:master2} or \ref{eq:master3}.
The overall objective function can be restated as 
\begin{equation}
h(\mathbf{GHI}) = \sum_{t=1}^T h_t(\mathrm{GHI}_t)
\end{equation}	
where $T$ is the total number of observations. 
Derivative-free optimization algorithms such as genetic algorithms, the Nelder Mead’s simplex method and particle swarm optimization algorithms are badly affected by increasing numbers of decision variables \cite{Pham2011,Rios2013}.
General purpose nonlinear solvers usually rely on calculating the objective function derivatives for all the values of the decision function. This means that $\frac{\partial h(\mathrm{GHI}_i) }{\mathrm{GHI}_j}$ must be calculated at each step. Even if it is possible to specify a pattern for the Hessian matrix to the trust-region-reflective algorithm in Matlab, which would significantly speed up the optimization, this algorithm requires the analytical gradient for $h$, which we do not possess \cite{Yuan2000}.\\
For this reason, we implemented a solver  for our problem. A comparison between our solver and fmincon computational time, when fmincon solves \ref{eq:master3} for each time step individually, is shown in table \ref{tab:solution_times}. The results are related to 1500 points. The relative difference in the solutions was below 1$\%$ with associated standard deviation of 2.02e-1 $\%$. 
Our algorithm took approximately 15.7 minutes to process one year of data with a temporal resolution of 10 minutes, on an Intel Xenon CPU E5-2697 v2 @ 2.70 GHz with 32.0 GB of RAM.

\begin{table}
	\renewcommand{\arraystretch}{1.3}
	\caption{Computational times comparison}
	\label{table_example}
	\centering
	\begin{tabular}{|c|c|c|}
		\hline 
		& mean s$/$sample & std s$/$sample \\ 
		\hline 
		fmincon & 2.32 & 3.1e-2 \\ 
		\hline 
		our solver & 1.8e-2 & 8.5e-3 \\ 
		\hline 
	\end{tabular}\label{tab:solution_times}
\end{table}

Since our objective function $h$ is in the form described in \ref{eq:decomp}, our solver simply minimizes $h(x_i) \quad \forall i \in [1,T]$ with a steepest-descent solution strategy. Since $h(GHI_i)$ could present local minima as previously stated in \ref{sec:methodology}, we initialized the solution with a grid search over the possible values of $\mathrm{GHI}$, as shown in the pseudocode \ref{alg:1}.\\
For each time step, we searched in a discrete space of possible values of $\mathrm{GHI}$, uniformly sampled from 0 to 
  
\begin{equation}
\mathrm{GHI}_{max,t} = k\mathrm{GHI}_{clear,t} 
\end{equation} 
where $\mathrm{GHI}_{clear,t}$ is the $\mathrm{GHI}$ calculated from the clear sky model, and $k$ is a safety factor accounting for the fact that particular cloud configurations can increase the measured GHI to above the clear sky values \cite{Inman2016}. 
We use a 30-step discretization for the grid search. Considering a standard irradiation of 1000 W/m2, this would result in an approximate accuracy level of 33.3 W/m2. 
We obtain  a set of $n_g$ guess vectors for $\mathrm{GHI}$, and for each of them we calculate the proxies and assess the hypothetic power produced by each PV plant (line 3 and 4 of algorithm \ref{alg:1}), and the PV estimation error matrix $\mathbf{E}_g  \in {\rm I\!R}^{T \times n_{pv}} $  (line 5).   
Then, for each time $t$ we find the best guess $\mathrm{GHI}_t^*$, which is the one that minimizes the average PV estimation error among the different PV plants (line 8-11 of algorithm \ref{alg:1}).
Note that the inner minimization of line 9 is inexpensive, since it consists of finding the position of the minimum element of the average of  $E_{g,t,i}$ over g, at a given $t$, where $i$ refers to the ith PV plant.\\
Once the initial guess for $\mathrm{GHI}$ has been obtained, the solution is refined as shown in the pseudocode \ref{alg:2}.
Starting from the first guess solution, we once again determine the PV estimation error at iteration $\nu$, and then calculate the gradient of the objective function introduced in \ref{eq:master3} with respect to $\mathrm{GHI}$:

\begin{equation}
\nabla_{\mathbf{GHI}} \mathbf{h} = \frac{1}{N}  \sum_{i=1}^{N} \mathcal{I}(\mathbf{E}) \mathcal{F}_{i,t}  \nabla_{\mathbf{GHI}} \mathbf{e}_{i}
\end{equation} 
where 
\begin{equation}
\nabla_{\mathbf{GHI}} \mathbf{e}_i = -\left(\nabla_{\mathbf{GHI}} \mathbf{Pr}\right)\boldsymbol{\omega}_i
\end{equation} 
note again that, since the function is time-separable, the only non-zero elements of $\nabla_{\mathbf{GHI}} \mathbf{Pr}$ are those related to observations occurring at the same time-step 
\begin{equation}
\frac{\partial \mathrm{Pr}_{i,j}}{\partial \mathrm{GHI}_k} = 0  \qquad \forall \quad j \not = k
\end{equation}
For this reason, the resulting tensor can be rewritten in matrix form such that  $\nabla_{\mathbf{GHI}} \mathbf{Pr} \in {\rm I\!R}^{T,n_p}$.
Each element of  $\nabla_{\mathbf{GHI}} \mathbf{Pr}$ is calculated as 

\begin{equation}
\frac{\partial \mathrm{Pr}_{i,j}}{\partial \mathrm{GHI}_i} = \frac{\mathrm{Pr}_{i,j}^+-\mathrm{Pr}_{i,j}}{\delta_{ghi}}
\end{equation}
where $Pr_{i,j}^+$ is the jth proxy at ith observation computed from $\mathrm{GHI}^+ = \mathrm{GHI}+\delta_{ghi}$.
\begin{algorithm}
	\caption{Initialize GHI}\label{alg:1}
	\begin{algorithmic}[1]
		\For{$g \in [1,n_{g}]$} \Comment{grid search initialization}
		\State $\mathbf{GHI}_g = g/n_g\mathbf{GHI}_{max} $ \Comment{linear rescale}
		\State $\mathbf{Pr}_g \gets \mathbf{GHI}_g$
		\State $\widehat{\mathbf{P}}_g =  \mathbf{Pr}_i\mathbf{\Omega}$
		\State $\mathbf{E}_g = \mathbf{P} -\widehat{\mathbf{P}}_g$
		\EndFor
		\State Find $GHI_{g}$ that minimize the mean PV error at each t
		\For{$t \in [1,T]$} 
		\State	$ g^\star \gets \argmin{g} \frac{1}{n_p} \sum_{p=1}^{n_p} E_{g,t,i}$ 
		\State	$\mathrm{GHI}^\star_t \gets \mathrm{GHI}_{g^\star,t}$ 
		\EndFor  
	\end{algorithmic}
\end{algorithm}

\begin{algorithm}
	\caption{Estimate GHI}\label{alg:2}
	\begin{algorithmic}[1]
		\While{$err_\nu \leq err_{\nu-1} \ \mathbf{and} \ \nu < \nu_{max}$}
		\State $\mathbf{Pr}_\nu \gets \mathbf{GHI}_\nu$
		\State $\widehat{\mathbf{P}}_\nu \gets  \mathbf{Pr}_\nu\mathbf{\Omega}$
		\State $\mathbf{E}_\nu \gets \mathbf{P} -\widehat{\mathbf{P}}_\nu$ \Comment{PV estimation error}
		\State $ \nabla_{\mathbf{GHI}} \mathbf{h}_{\nu}\gets \mathbf{GHI}_{\nu}+\delta_{ghi}, \mathcal{I}(\mathbf{E}), \mathcal{F}_{i,t}$ 
		\State $err_{\nu} = \norm{\mathbf{E}}_{fr}$
		\For{$t \in [1,T]$}
		\State $ \epsilon_{t,\nu} = \frac{1}{N}\sum_{i=1}^{N}\vert(\mathbf{e}_{i,\nu}-\mathbf{e}_{i,\nu-1}) \mathcal{I}(\mathbf{E})\mathcal{F}_{i,t} \vert \leq 0 $
		\If{$ \thicksim \epsilon_{t,\nu} $}
		\State ${\lambda}_{t,\nu+1} = k{\lambda}_{t,\nu}$
		\State ${\lambda}_{t,\nu} = 0$
		\EndIf
		\EndFor
		
		\State $\mathbf{GHI}_{\nu+1} = \mathbf{GHI}_{\nu} - \boldsymbol{\lambda}_{\nu} \nabla_{\mathbf{GHI}} \mathbf{h}_{\nu}$
		\EndWhile\label{euclidendwhile}
		
	\end{algorithmic}
\end{algorithm}
Lines 7 to 13 in algorithm \ref{alg:2} describe the $\boldsymbol{\lambda}_\nu$ update strategy, where $\boldsymbol{\lambda}_\nu$ is a vector of coefficients describing how much GHI must be shifted in the direction of the objective function gradient, at the $\nu$ iteration. Instead of using a backtracking strategy, which performs a line search on parameter $\lambda_{t,\nu}$, we applied an exponential decay on $\lambda_{t,\nu}$, in the attempt to reduce the total number of function evaluations. 
Since the objective function is not monotone in GHI, at each iteration $\nu$ we check  if the mean estimation error  has decreased. In this case, $\lambda_{t,\nu}$ is unchanged, otherwise $\lambda_{t,\nu}$ is set to zero (which results in not updating $\mathrm{GHI}_{\nu,t}$) and the new $\lambda_{t,\nu+1}$  decreases by a factor $k<1$.\\

\subsection{Trust function and outlier detection}
We now illustrate the method used to construct the trust function $\mathcal{F}(t)$ and the outlier detection function $\mathcal{I}(\mathbf{E})$.
As previously stated in Section \ref{sec:methodology}, $\mathcal{F}(t)$ weights the PV estimation differently based on the sun position. Greater importance is attached to signals with lower relative estimation errors during clear sky conditions for a given sun azimuth and elevation. Since we do not possess the real $\mathrm{GHI}$, clear sky periods must be estimated. In order to use as much data as possible for shadow detection, a different method from the one introduced in \ref{sec:orientation} is used. We estimate the relative PV estimation error under clear sky conditions as:
\begin{equation}\label{eq:clear_sky}
\mathrm{re}_{clear,i}(\boldsymbol{\gamma_s},\boldsymbol{\theta_z}) = \mathrm{Q}_{0.01,\boldsymbol{\gamma_s},\boldsymbol{\theta_z}}\left(\frac{\widehat{\mathbf{P}}_i(\mathbf{GHI}_{clear})-\mathbf{P}_i}{\mathbf{P}_i}\right)
\end{equation}   
where $\mathrm{Q}_{0,01}$ is the 1$\%$ quantile, $\boldsymbol{\gamma_s}$ and $\boldsymbol{\theta_z}$ are the sun azimuth and zenith angles, discretized with a 2$^\circ$ step.

\begin{figure}[h]
	\centering
	\includegraphics[width=1\linewidth]{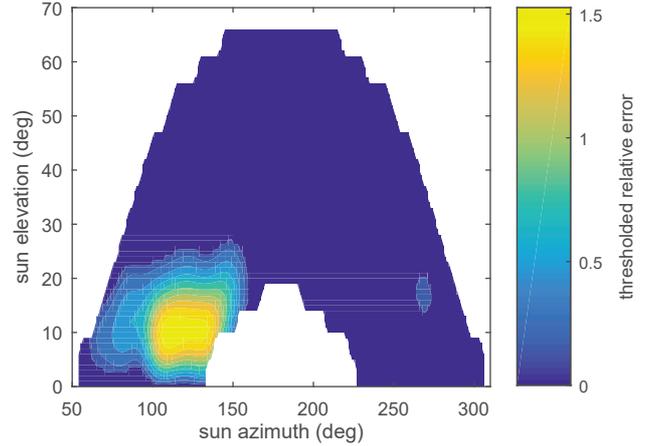}
	\caption{Thresholded smoothed estimated error as a function of sun position, for a given PV plant. This particular PV plant presents significant shadowing during morning hours}
	\label{fig:shadowerror}
\end{figure}

Since the $\mathrm{re}_{clear}(\gamma_{s,t},\theta_{z,t})$ map is assumed to be affected by noise, in order to obtain a more significant representation of the shadow pattern, we fit  a gaussian process on top of it. We then apply a threshold to eliminate the lowest values in the map, which could be due to the lack of observed clear sky conditions in the corresponding sun position.
An example of the resulting thresholded map for a given PV plant is shown in figure \ref{fig:shadowerror}. 

Once the PV estimation error has been established, we use it to map those signals that are more accurately calculated in a given combination of $\gamma_{s,t}$ and $\theta_{z,t}$, through an inverse relation:
\begin{equation}\label{eq:metrics}
d_{t,i} = re_{clear,i}(\gamma_{s,t},\theta_{z,t})^{-1}
\end{equation}
Finally, since we do not want the change in $\mathrm{GHI}$ to be unbounded , we normalize the obtained distances:
\begin{equation}\label{eq:trust_function}
\mathcal{F}_i(\gamma_{s,t},\theta_{z,t}) = \frac{d_{i,t}}{\sum_{i=1}^{N}d_{i,t}}
\end{equation}
where $d_i(t)$ is the distance of signal $i$ at time step $t$.
The second strategy for improving the $\mathrm{GHI}$ estimation accuracy is to detect outliers in different signals at each time-step $t$. Intuitively, if we possess more than one power signal, we can study the distribution of the various estimation errors and exclude from the objective function those signals that at time-step $t$ are labeled as outliers, applying a standard outlier detection method.
We used Tukey's outlier detection method \cite{Tukey1977}, which is based on the interquartile range, and which can be applied to non-symmetric data distributions. We can now define the outliers detection function $\mathcal{I}$ as:
\begin{equation}
\mathcal{I}(e_{i,t})= 
\begin{cases}
0,&\text{if} \ e_{i,t} \leq \mathrm{Q}_{0.25} -k_q\mathrm{IQ} \ \vee \ e_{i,t} \geq \mathrm{Q}_{0.75} +k_q\mathrm{IQ}\\
1,&\text{otherwise}
\end{cases}
\end{equation}
where $\mathrm{IQ}$ is the interquartile range: i.e. $\mathrm{Q}_{0.75}-\mathrm{Q}_{0.25}$ and $k_q$ is a parameter dependent on the assumed distribution of $e_{t}$. We used $k_q = 1.5$, which corresponds to considering approximately 1$\%$ of the points as outliers, under normal data distribution conditions.
When a point is identified as an outlier, it is not used to correct $\mathrm{GHI}$. This is done by setting to zero the corresponding elements of the objective function gradient. Formally:
\begin{equation}\label{eq:outlier_exclusion}
\nabla_{\mathbf{GHI}}\mathrm{h}_{\nu,i_o,t_o} = 0
\end{equation}
where $i_o,t_o$ are the signal and timestep marked as outliers.

\section{Evaluation on case studies}\label{sec:case}
The proposed method has been tested on two case studies with multiple power signals. Both case studies are located in Switzerland and are particularly challenging for $\mathrm{GHI}$ estimation since they present:
\begin{itemize}
	\item multiple PV orientations (even at single inverter level)
	\item different nominal powers for each PV installation
	\item significant shading, due to nearby objects and/or horizon profile
	\item partial PV curtailment
\end{itemize}

The first case study is located in Biel-Benken, on the northern Swiss plateau close to the German and French borders. It consists of 4 residential rooftop PV installations with nominal powers ranging from 6.6 to 10.7 kWp. The mean distance between the PV plants and the pyranometer is approximately 150 meters. The PV plants are affected by local shading, due to the presence of chimneys and nearby buildings. One PV plant has two different module orientations (mounted on two folds of the same roof). The effect of the horizon in this region is negligible.\\
The second case study is located in Lugano, a hilly region in the alpine foothills. In this case the power signals are related to 5 different industrial PV plants, with nominal power ranging from 126 to 275 kWp. Several inverters are installed in each plant, making a total number of 50 inverters. The mean distance between the PV plants and the pyranometer is approximately 300 meters. In this case the horizon is non-negligible, due to the presence of significant topographical relief formations.\\
For each case study, both on-ground measurements and satellite data are  used for performance assessment. 
At each location, an ISO 9060 secondary standard pyranometer (CMP10 and CMP21, Kipp\&Zonen, Delft, The Netherlands) is  used as ground-truth reference. In the first case-study, the pyranometer is  mounted on one of the roofs hosting the PV installations, while in the second case the pyranometer is  located in the SUPSI Trevano campus. 
The results were  also compared against two different satellite-based irradiation models: MACC-RAD and SICCS. MACC-RAD uses the Heliosat-4 method \cite{Thomas2016}, while SICCS is based on a Cloud Physical Properties model \cite{Greuell2013}. Both models are based on Meteosat satellite images. The MACC-RAD data are freely accessible, while SICCS data are sold by 3E.\\


The data for the Biel-Benken case study refer to the period from August 1$^\mathrm{st}$, 2015 to August 1$^\mathrm{st}$, 2016, with a 1-minute sampling time. Since this case presents a great annual variation in terms of shadow pattern, in order to increase the method accuracy, we repeatedly identified $\boldsymbol{\Omega}$ for different time splits of the data. Then, we chose $\boldsymbol{\Omega}$ as the one that minimizes the total RMSE on the mean PV estimation error. In particular, table \ref{data_split} shows the attempted split and the achieved RMSE for the PV estimation error and for the GHI estimation error, reported here only for comparison. The chosen split uses 3 months data folds .
\begin{table}
	\renewcommand{\arraystretch}{1.3}
	\caption{Data splits for Biel-Benken $\boldsymbol{\Omega}$ identification }
	\label{data_split}
	\centering
	\begin{tabular}{|c|c|c|c|c|c|}
		\hline 
		split [days] & 365 & 182  & 121  & $\mathbf{91}$ & 73 \\ 
		\hline 
		PV$_{RMSE}$ [-]  &6.6e-2 & 5.7e-2 & 5.64e-2  & $\textbf{5.6e-2}$   & 6.4e-2 \\ 
		\hline 
		GHI$_{RMSE}$ [W/m2] &41.7 & 34.8 & 34.4  & $\mathbf{33.5}$  & 35\\ 
		\hline 
	\end{tabular}
\end{table}

\begin{figure*}[h!]
	\centering
	\includegraphics[width=1\textwidth]{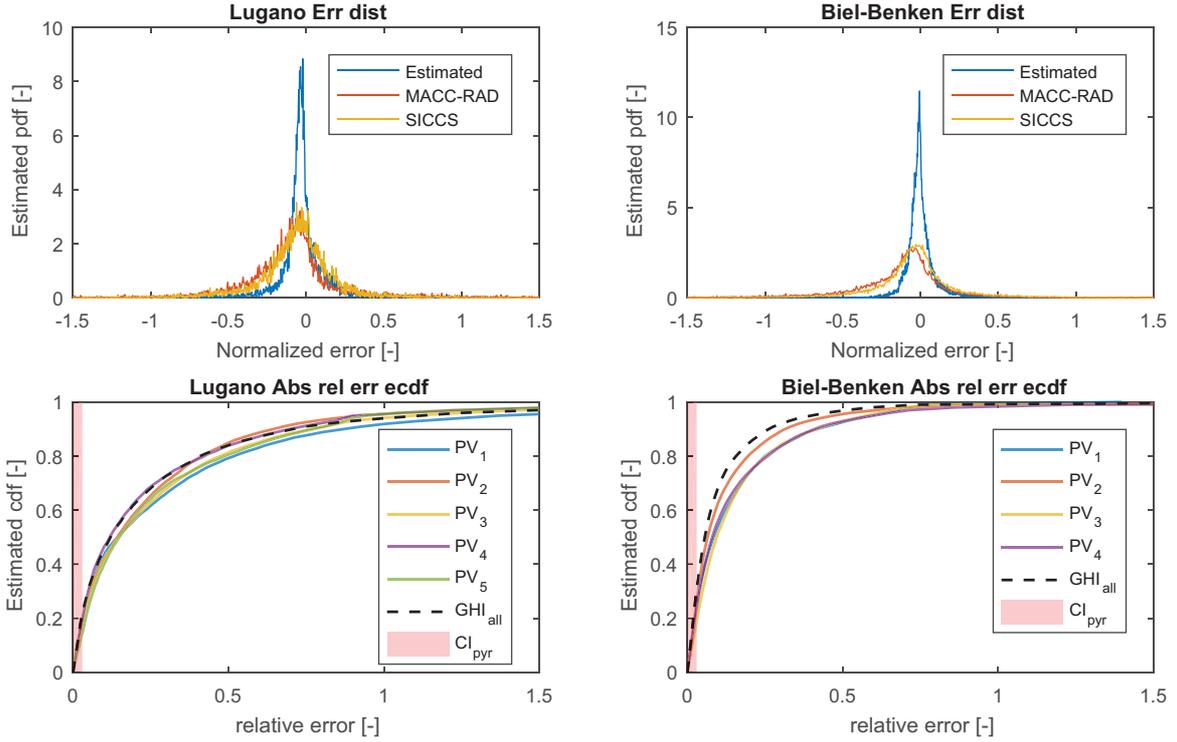}
	\caption{Error distribution for the two case studies with sampling period of 10 minutes. Upper part: empirical probability density functions of the normalized error distributions for the estimated $\mathrm{GHI}$ signal and for the $\mathrm{GHI}$ signal from MACC-RAD and SICCS models. Lower part: empirical cumulated density functions for the absolute relative errors of the estimated signals. Colored lines refers to the $\mathrm{GHI}$ signal estimated using single PV plants power signals, the black dashed line refers to the $\mathrm{GHI}$ signal estimated with power signals of all the PV plants. Pale red bands refers to the confidence interval of the pyranometer}
	\label{fig:errordists}
\end{figure*} 

The data for the Lugano case study refer to the period from January 1$^\mathrm{st}$, 2016 to June 1$^\mathrm{st}$, 2016, with a 10-minute sampling time.
In both case studies, equation  \ref{eq:master} is solved for each PV plant separately, obtaining 4 and 5 different GHI estimations, respectively. 
Equation \ref{eq:master3} is then solved using all the signals from the different PV plants, in an attempt to improve the GHI estimation.\\ 
Figure \ref{fig:errordists} summarizes the main results of two case studies, for a sampling time of 10 minutes. The top graphs show the normalized error distributions between $\mathrm{GHI}_{py}$ and the two satellite-based models and between $\mathrm{GHI}_{py}$ and the solution of problem \ref{eq:master3}, where $\mathrm{GHI}_{py}$ is the signal measured by the pyranometers. The normalization is obtained by dividing the error distribution by  a constant value $k_n$, defined as the mean value of non-zero $\mathrm{GHI}$ observations:\\

\begin{equation}
k_n = \frac{1}{T^*}\sum_{t=1}^{T^*} \mathrm{GHI}_{py,t} \qquad \forall \  \mathrm{GHI}_{py,t} >0
\end{equation}

where $T^*$ is the number of non-zero elements of vector $\mathbf{GHI}_{py}$.
In both case studies, estimating GHI from on-ground measurements significantly narrows the error distribution.
The lower part of figure \ref{fig:errordists} shows the estimated probability density function for the absolute relative errors. Each line represents the GHI calculated using one single PV plant, while the thick slashed line is related to the GHI calculated using all the PV signals. The red band represents the typical pyranometer level of accuracy.
The results suggest that using more than one signal increases the robustness and accuracy of the method.  This is partly due to the trust functions and outlier detection function, as better explained in figure \ref{fig:rmsebars}.
As we can see, the normalized root mean squared error decreases when we remove erratic observations from the objective function, and when we use the trust function to weight the signals. \\
Figure \ref{fig:errordists} shows that the proposed method has low bias with respect to the satellite-based methods. In order to gain additional insight into the error distributions, we performed a bias-variance error decomposition:

\begin{figure*}[h!]
	\centering
	\includegraphics[width=1\textwidth]{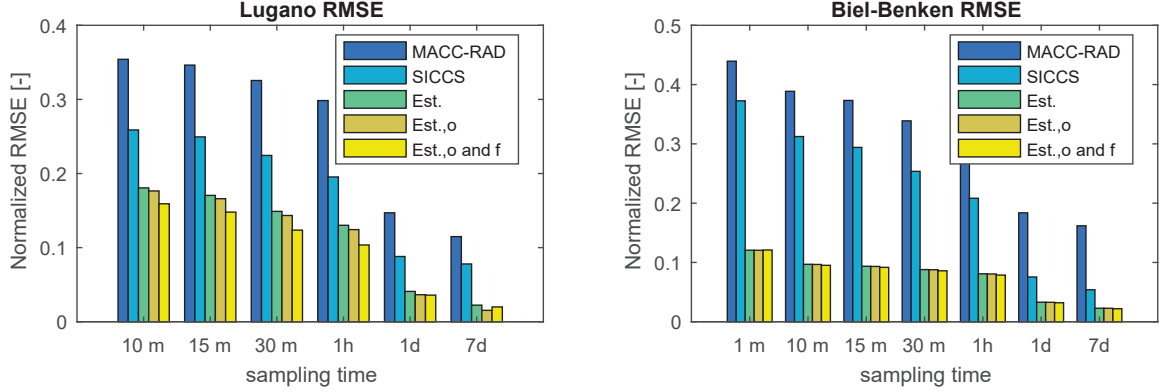}
	\caption{Normalized RMSE for the two test cases, for different aggregation times. The blue and light blue bars represent the RMSE of the MACC-RAD and SICCS models. The green, brown and yellow bars represent the RMSE of the estimated GHI using our method without corrections, with the outliers detection function and with both outliers detection and trust functions, respectively.}
	\label{fig:rmsebars}
\end{figure*}

\begin{figure*}[h!]
	\centering
	\includegraphics[width=1\linewidth]{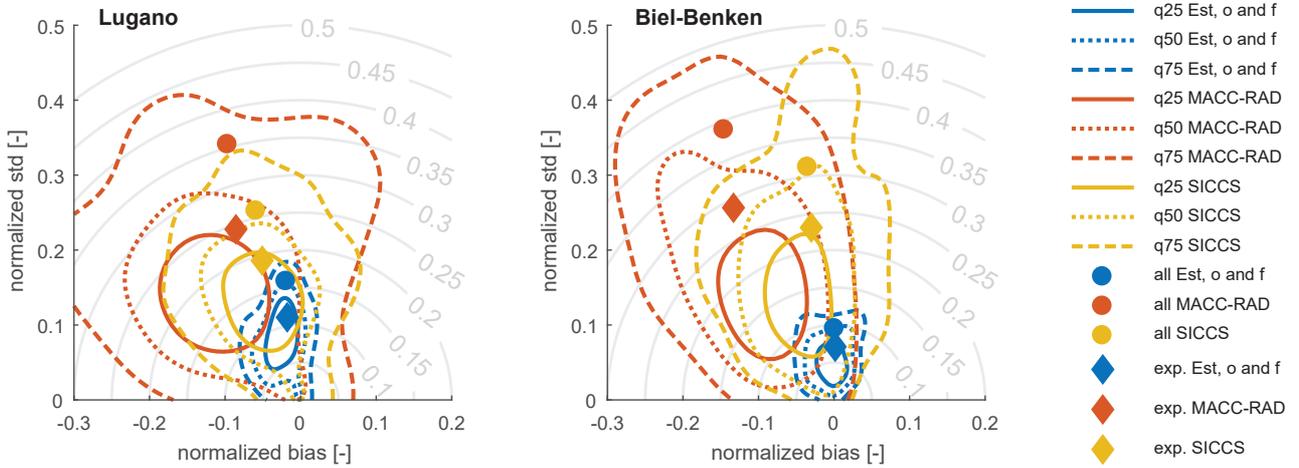}
	\caption{Normalized bias and standard deviation of the MACC-RAD and SICCS services and the proposed method (using outlier detection and trust function), for the two test sites. Colored lines are estimated regions containing 25, 50 and 75 $\%$ of the observations, when considering daily subsets. Filled circles represent normalized squared bias and normalized standard deviation for the whole datasets, while diamonds represent the expected daily normalized squared bias and normalized standard deviation. Grey lines show contour lines for the normalized RMSE.}
	\label{fig:kdesm}
\end{figure*}

\begin{align}
RMSE^2_{\mathcal{D}} =& \left({\rm I\!E}_{\mathcal{D}}e\right)^2 + \left[{\rm I\!E}_{\mathcal{D}}\left(e-{\rm I\!E}_{\mathcal{D}}e \right)^2\right]^2\\
=& \ \mathrm{bias}^2_{\mathcal{D}} + \mathrm{std}^2_{\mathcal{D}}
\end{align}

where $e$ is the estimation error, $\mathcal{D}$ is a given dataset and ${\rm I\!E}_{\mathcal{D}}$ is the expectation over  the dataset $\mathcal{D}$.
We calculated $\mathrm{bias}_{\mathcal{D}}$ and $\mathrm{std}_{\mathcal{D}}$ using daily datasets. We performed the calculation using the maximum available time resolution, for our method with trust function and outlier detection and for the two satellite-based methods. This procedure generates bivariate distributions in terms of bias and standard deviation. For visualization reasons, instead of showing all the points generated by this daily decomposition, estimations of the regions containing 25, 50 and 75 $\%$ of the points, respectively, are plotted in figure \ref{fig:kdesm}. These estimations were obtained using the kernel density smoother Matlab function $\texttt{ksdensity}$.
Normalized bias and normalized standard deviation for all the observations in the datasets are shown (filled circles).
We also show the daily expected values for the normalized bias and normalized standard deviation (diamonds).
We can see from figure \ref{fig:kdesm} that, in comparison with other methods, the proposed method has a narrower distribution in terms of bias and standard deviation. 

For the Biel-Benken case study, the exact tilt, azimuth and nominal power of the installed PV plants are known. In order to check if these values were estimated correctly, we compared the ground truth with the identified $\boldsymbol{\Omega}$, plotted as a function of azimuth and elevation in figure \ref{fig:identifiedProxies}. 
It can be seen that the non-zero coefficients of $\boldsymbol{\Omega}$ are close to those of the real PV plants. In fact, except for the second PV plant, whose PV panels present more than one orientation, the real orientations lie in the convex-hull formed by the non-zero coefficients.


\section{Conclusions}\label{sec:conclusions}
In this paper, we present an unsupervised method for estimating global horizontal irradiance from the AC measurements of one or more PV plants, consisting of PV modules of unknown nominal power and orientation. The only inputs to the method are the AC power signals from the PV plants, with corresponding timestamp, and their approximate location in terms of latitude, longitude and altitude.\\
An algorithm was developed to speed up the optimization of the underlying non-linear non-convex problem. In terms of computational time, this compares favorably with existing general-purpose solvers.\\

\begin{figure}[h]
	\centering
	\includegraphics[width=1\linewidth]{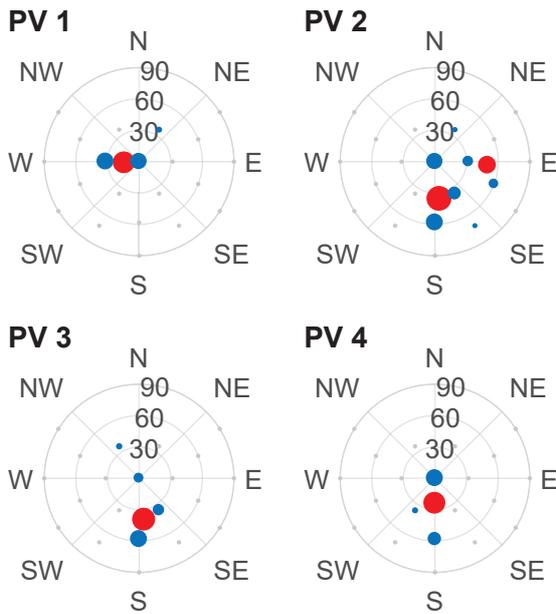}
	\caption{Identified $\boldsymbol{\Omega}$ for the Biel-Benken case study. The grey dots indicate the orientations of the proxies, each dot referring to a different column in the $\mathbf{Pr}$ matrix. The blue dots represent the identified coefficients of $\boldsymbol{\Omega}$ while red dots represent the ground truth. The size of the dots is proportional to the identified/nominal power for a given orientation.}
	\label{fig:identifiedProxies}
\end{figure}

The method was tested in two different case-studies, both presenting shading and partial curtailment. With respect to other existing satellite-based methods, the results show a significant improvement in the GHI estimation, in terms of RMSE, as shown in figure \ref{fig:rmsebars}. In both case studies, the relative calculation error is within the secondary standard pyranometer confidence interval for roughly 20-30$\%$ of the observations, as shown in \ref{fig:errordists}.
The method can correctly identify the orientation and nominal power of the PV modules, even when the PV plant presents PV fields with different orientations. See figure \ref{fig:identifiedProxies}. \\
The method relies on constructing proxies for the electrical power output of the PV modules. This depends on a set of parameters, namely $\beta$, $\gamma$, $k_1$, $k_2$, $k_3$, $k_4$, which in this study were kept fixed. Further work is required in order to determine the influence of these parameters on the performance of the algorithm in terms of estimation accuracy.
In this work the Maxwell empirical disc model has been used to disaggregate the direct and diffuse irradiance. In recent years, other models have been developed, which have been shown to provide better results \cite{Gueymard2016}, as for example the ENGERER2 model \cite{Engerer2015}. In future work we will study the effect of different separation models on the algorithm performance.\\
The proposed method will be used in future studies, in order to disaggregate PV generation from electricity demand, in an attempt to increase the accuracy of aggregated net load short-term forecasts in a low voltage grid. The developed algorithm is freely available as open-source code at\\
 \url{https://github.com/supsi-dacd-isaac/GHIEstimator.}

\section*{Acknowledgments}
The authors would like to thank CTI - Commission for Technology and Innovation (CH), and SCCER-FURIES - Swiss Competence Center for Energy Research - Future Swiss Electrical Infrastructure, for their financial and technical support to the research work presented in this paper.


\end{document}